\documentclass[letterpaper, 10 pt, conference]{ieeeconf}
\IEEEoverridecommandlockouts
\usepackage{graphicx}
\usepackage{amsmath}
\usepackage{amssymb}
\usepackage{booktabs}
\usepackage{multirow}
\usepackage{xcolor}
\usepackage{cite}
\usepackage[ruled,vlined,linesnumbered]{algorithm2e}
\usepackage[hidelinks]{hyperref}

\title{\LARGE \bf
Verify2Act: Critic-Guided Latent World Models for Verifying Language-Conditioned Manipulation Plans
}

\author{Chrisantus Eze$^{1}$ and Christopher Crick$^{1}$
\thanks{$^{1}$The authors are with the Department of Computer Science, Oklahoma State University, Stillwater, OK, USA.
        {\tt\small chrisantus.eze@okstate.edu, chriscrick@cs.okstate.edu}}%
}

\begin{document}
\maketitle
\thispagestyle{empty}
\pagestyle{empty}

\begin{abstract}
Vision-Language Models (VLMs) can propose multi-step manipulation plans, but they have no grounded model of how a scene will change, and the pixel-level world models that could check those plans are slow and drift over long horizons. We present Verify2Act, a modular verification framework that separates semantic action proposal from latent physical verification. A VLM proposes candidate plans, a latent world model (V2A-WM) imagines each of them in the DINOv2 feature space, and a dual-head critic checks the imagined rollouts for temporal consistency and goal satisfaction before anything is executed. V2A-WM extends recent flow-matching world models with cross-attention action grounding and a causal temporal history window. We deploy the full system on a low-cost robot arm, training the world model and critic only on data generated in a digital twin. Across six language-conditioned tasks, Verify2Act reaches 83.3\% task success, compared with 61.7\% for a latent world model without our architectural changes, 53.3\% for the VLM alone, and 46.7\% for a pixel-diffusion world model. It verified every plan it executed and incurred no planning-induced failures: all of its failed episodes were caused by low-level grasp or placement slips. In simulation, Verify2Act obtains the highest five-task chain success on CALVIN among the compared baselines with 30$\times$ fewer reflective VLM calls than pixel-diffusion verification, and the highest success on cluttered nut assembly.
\end{abstract}


\section{Introduction}
\label{sec:intro}

Solving multi-stage manipulation requires both semantic planning and physical verification. Vision-language models (VLMs) are strong at semantic planning ~\cite{ahn2022can, driess2023palm, brohan2023rt}, but lack a grounded model of how actions change the scene. Pixel-space or generative world models can provide this foresight ~\cite{du2023learning, gu2026say, feng2025reflective, black2024zero}, but are computationally expensive and prone to visual drift over long horizons ~\cite{zhang2026learning}.

Latent visual world models offer a more efficient alternative. DINO-WM~\cite{zhou2024dino} and RLA-WM~\cite{zhang2026learning} model dynamics in DINOv2~\cite{oquab2023dinov2} feature space, but neither is designed for language-driven plan verification: DINO-WM searches with CEM, while RLA-WM uses a Markovian architecture without temporal memory, limiting language-conditioned planning, robustness to occlusion, and accumulating background drift over multi-step rollouts.

We present \textbf{Verify2Act}, a modular verification framework that decouples semantic action proposal from latent physical verification (Fig.~\ref{fig:architecture}). The VLM serves only as a semantic heuristic that proposes a small set of candidate plans. These candidates are then imagined by V2A-WM, which builds on the flow-matching cores of DINO-WM and RLA-WM with two changes: (1) \textbf{cross-attention action grounding} between DINOv2 patch features and CLIP~\cite{radford2021clip} tokens, which ties language commands to visual object patches; and (2) a \textbf{causal temporal history} window with validity masks, which preserves visual memory across gripper occlusions. A dual-head critic then scores the imagined rollouts for temporal consistency and goal satisfaction before any action is committed.

We evaluate Verify2Act on a robot arm and in simulation. For the robot, the world model and the critic are trained only on data generated in a digital twin of the workspace, and an optional real2sim input-alignment step carries them over to the real camera. Our contributions are:
\begin{itemize}
    \item \textbf{Verify2Act}, a low-level-policy-agnostic verification layer that decouples semantic action proposal from latent physical simulation.
    \item \textbf{V2A-WM}, a visual feature-based flow-matching world model with cross-attention action grounding and causal temporal history.
    \item A dual-head \textbf{critic} that scores temporal consistency and goal satisfaction for beam-search plan selection.
    \item An \textbf{evaluation} of four verification strategies on a robot arm over six language-conditioned tasks, with models trained only on digital-twin data and with planning errors separated from execution errors, together with simulation results on CALVIN~\cite{mees2022calvin} and Robosuite~\cite{zhu2020robosuite}.
\end{itemize}

\section{Related Work}
\label{sec:related}
\textbf{VLMs and Foresight in Robot Planning.}
Reactive VLM planners such as SayCan~\cite{ahn2022can}, RT-2~\cite{brohan2023rt}, and PaLM-E~\cite{driess2023palm} map observations directly to actions without modeling the physical consequences of those actions. To introduce foresight, recent work has paired VLMs with pixel-space diffusion models that synthesize future visual states~\cite{feng2025reflective, black2024zero, du2023learning, gu2026say, jia2025video2act, pai2025mimic}, or folded semantic planning and visual prediction into a single backbone~\cite{li2025unified, zhu2025unified, kim2026cosmos}. ReflectVLM~\cite{feng2025reflective}, for example, uses diffusion-generated imagined states to iteratively revise plans. Similarly, FlowMPC~\cite{flowmpc2024} applies world models at test time to guide policy trajectories. Across all these approaches, however, raw pixel prediction remains computationally expensive and prone to accumulating visual artifacts over long planning horizons~\cite{zhang2026learning}. Verify2Act avoids pixel generation entirely by performing verification in the latent feature space.

\textbf{Visual Feature-Based World Models.}
Modeling dynamics over pretrained visual features avoids pixel reconstruction while retaining semantic and geometric information~\cite{hou2026world, hafner2023dreamerv3, yang2023learning}. DINO-WM~\cite{zhou2024dino} predicts DINOv2 features and plans with CEM, while RLA-WM~\cite{zhang2026learning} uses residual latent actions with flow matching. Other recent work explores flow-based feature dynamics ~\cite{flowwm2024}, latent modulation~\cite{mowm2024}, JEPA-style prediction~\cite{assran2025vjepa}, and latent VLA models~\cite{zheng2025flare, sun2026vlajepa, miao2026jepavla, su2026wog, chen2026dial}. Verify2Act builds on these ideas by introducing language-to-patch cross-attention and causal temporal history for plan verification.

\textbf{Hierarchical and Sampling-Based Planning.}
Task and Motion Planning (TAMP)~\cite{garrett2020tamp} decomposes manipulation into symbolic plans and low-level motion primitives, while data-driven approaches such as Points2Plans~\cite{huang2025points2plans} learn relational dynamics from point clouds. Verify2Act instead performs proposal and verification directly in visual feature space, without task-specific predicate classifiers.

\section{Method}
\label{sec:method}

\begin{figure*}[t]
    \centering
    \includegraphics[width=0.75\textwidth]{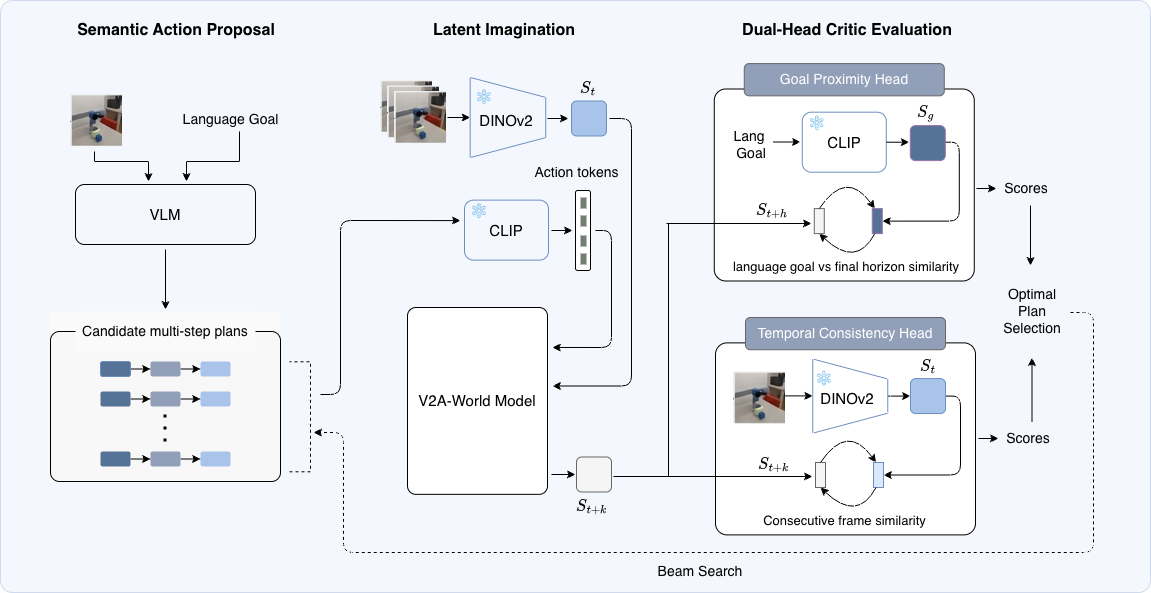}
    \caption{\textbf{Verify2Act overview.} \textbf{Stage 1:} a VLM proposes candidate multi-step plans from the camera observation and the language instruction. \textbf{Stage 2:} V2A-WM rolls out each candidate in the frozen DINOv2 latent space ($S_t \rightarrow S_{t+k}$), conditioned on CLIP action tokens. \textbf{Stage 3:} a goal head scores the terminal latent state against the goal, and a temporal consistency head checks consecutive imagined states; the best verified plan is executed.}
    \label{fig:architecture}
\end{figure*}

\subsection{Problem Formulation and Overview}
\label{sec:method_overview}
The agent receives an RGB observation $o_t$ and must execute a sequence of $H$ primitive actions $\boldsymbol{a}_{1:H} \in \mathcal{A}^H$ (e.g., \texttt{pick}, \texttt{place}, \texttt{insert}) to satisfy a natural-language goal $g$. We learn a latent dynamics model $\hat{\mathcal{T}}$ over DINOv2 patch features $S_t = E_{\text{DINO}}(o_t)$. Given a predicted rollout $\hat{S}_{1:H}$ under actions $\boldsymbol{a}_{1:H}$ (with $\hat{S}_0 = S_t$ and $\hat{S}_k = \hat{\mathcal{T}}(\hat{S}_{k-1}, a_k)$), the planning objective is
\begin{equation}
    \boldsymbol{a}_{1:H}^* = \arg\max_{\boldsymbol{a}_{1:H} \in \mathcal{A}^H} \mathcal{C}\big(\hat{S}_{1:H},\; g\big),
\end{equation}
where $\mathcal{C}$ scores the trajectory against the goal. Verify2Act makes this objective tractable through a three-stage loop (Fig.~\ref{fig:architecture}, Algorithm~\ref{alg:v2a}): the VLM proposes candidates, V2A-WM rolls them out in latent space, and the critic scores and filters them step-wise.

\begin{algorithm}[t]
\footnotesize
\SetAlgoLined
\DontPrintSemicolon
\caption{Verify2Act: Parallel Trajectory Planning}
\label{alg:v2a}
\KwIn{Observation $o_t$, goal $g$, beam width $B$, horizon $H$, thresholds $\theta_c, \theta_p, \theta_{\mathrm{conf}}$, re-plan budget $R_{\max}$}
\KwOut{Best action plan $\boldsymbol{a}^*_{1:H}$}

\tcp{Stage 1 — Propose}
$\mathcal{P} \leftarrow \{\boldsymbol{a}^{(1)}, \ldots, \boldsymbol{a}^{(B)}\} \leftarrow \mathrm{VLM.propose}(o_t,\, g,\, B)$\;
$\boldsymbol{a}^* \leftarrow \boldsymbol{a}^{(1)}$;\quad $\text{best\_score} \leftarrow -\infty$\;

\For{$r = 0$ \KwTo $R_{\max}$}{%
    \tcp{Stage 2\,\&\,3 — Simulate and Verify each candidate}
    \For{$\boldsymbol{a}^{(i)} \in \mathcal{P}$}{%
        $\hat{S}_0 \leftarrow E_{\mathrm{DINO}}(o_t)$;\quad $\bar{s}_{\mathrm{tc}} \leftarrow 0$;\quad $\text{ok} \leftarrow \texttt{true}$\;
        \For{$k = 1$ \KwTo $H$}{%
            $\hat{S}_k \leftarrow \mathrm{WorldModel}(\hat{S}_{k-1},\, a^{(i)}_k,\, \mathcal{H}_t)$\;
            $(s_k,\, \hat{\sigma}_k) \leftarrow \mathrm{Critic}_{\mathrm{TC}}(\hat{S}_{k-1},\, \hat{S}_k)$\;
            \lIf{$\hat{\sigma}_k \geq \theta_{\mathrm{conf}}$ \textbf{or} $s_k < \theta_c$}{%
                $\text{ok} \leftarrow \texttt{false}$;\quad \textbf{break}\quad\tcp*[f]{incoherent step}
            }
            $\bar{s}_{\mathrm{tc}} \leftarrow \bar{s}_{\mathrm{tc}} + s_k / H$\;
        }
        \If{$\text{ok}$}{%
            $s_{\mathrm{gp}} \leftarrow \mathrm{Critic}_{\mathrm{GP}}(\hat{S}_H,\, g)$\;
            $\mathcal{C}^{(i)} \leftarrow \alpha\,s_{\mathrm{gp}} + (1-\alpha)\,\bar{s}_{\mathrm{tc}}$\;
            \lIf{$\mathcal{C}^{(i)} > \text{best\_score}$}{%
                $\boldsymbol{a}^* \leftarrow \boldsymbol{a}^{(i)}$;\quad $\text{best\_score} \leftarrow \mathcal{C}^{(i)}$
            }
        }
    }
    \lIf{$\text{best\_score} \geq \theta_p$}{\Return{$\boldsymbol{a}^*$}\quad\tcp*[f]{verified}}
    \tcp{Reflect and replan if no candidate passes}
    $\mathcal{P} \leftarrow \mathrm{VLM.reflect}(o_t,\, g,\, \boldsymbol{a}^*)$\;
}
\Return{$\boldsymbol{a}^*$}\tcp*[f]{budget spent: best unverified plan}
\end{algorithm}

\subsection{Semantic Action Proposal}
\label{sec:method_vlm}
At each planning step, the VLM receives the current RGB observation, goal, and action history and produces $B$ diverse $H$-step candidate plans in one few-shot call. It is used only to narrow the combinatorial search space; physical feasibility is assessed by V2A-WM and the critic.

\subsection{Visual Feature-Based World Model}
\label{sec:method_world_model}

\begin{figure*}[t]
    \centering
    \includegraphics[width=0.70\textwidth]{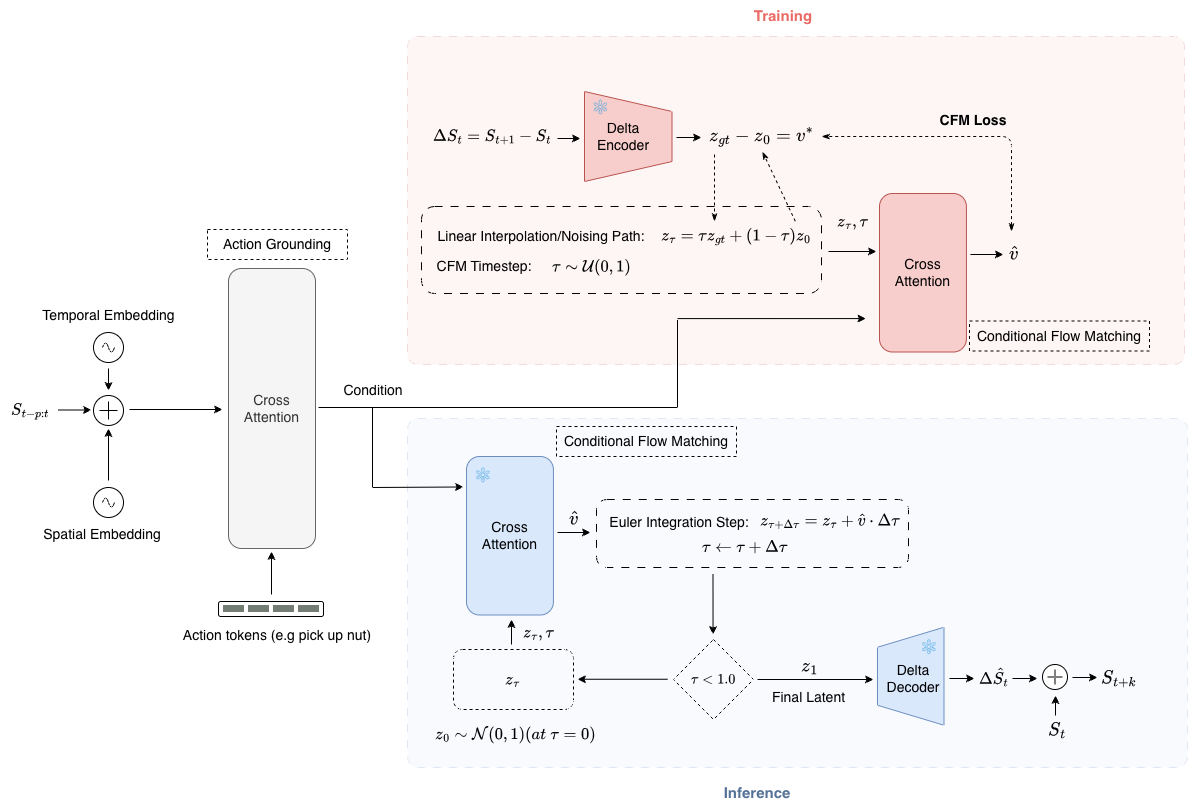}
    \caption{\textbf{V2A-WM training and inference.}
    \textbf{Training:} the delta encoder gives the target latent $z_{gt} = f_{\text{enc}}(\Delta S_t)$; we interpolate $z_\tau = \tau z_{gt} + (1-\tau)z_0$ and the flow-matching transformer, conditioned on past frames and CLIP action tokens through cross-attention, predicts the velocity $\hat{v}$.
    \textbf{Inference:} a single Euler step integrates from noise $z_0$ to $z_1 = z_0 + \hat{v}$, and the delta decoder reconstructs $\hat{S}_{t+1} = S_t + f_{\text{dec}}(z_1)$.}
    \label{fig:world_model}
\end{figure*}

V2A-WM operates on patch features $S_t \in \mathbb{R}^{P \times D}$ from a frozen DINOv2 backbone and, following RLA-WM~\cite{zhang2026learning}, trains in two stages (Fig.~\ref{fig:world_model}). First, a bottleneck autoencoder compresses the inter-frame residual $\Delta S_t = S_{t+1} - S_t$ into latent tokens $z = f_{\text{enc}}(\Delta S_t) \in \mathbb{R}^{M \times D_z}$ ($M \ll P$) under the reconstruction loss $\|\Delta S_t - f_{\text{dec}}(f_{\text{enc}}(\Delta S_t))\|^2$; both are then frozen. Second, a conditional flow-matching network~\cite{lipman2022flow} $v_\theta$ predicts the residual latent conditioned on the action $a_t$ and the temporal history $\mathcal{H}_t$: $\hat{v} = v_\theta(z_\tau, \tau;\, \mathcal{H}_t, a_t)$. With flow time $\tau \in [0, 1]$ and noise $z_0 \sim \mathcal{N}(0, I)$, we use the linear path $z_\tau = \tau z_{gt} + (1 - \tau) z_0$ and minimize
\begin{equation}
    \mathcal{L}_{\text{FM}} = \mathbb{E}_{\tau, z_0, z_{gt}}\left\| \hat{v} - (z_{gt} - z_0) \right\|^2 .
\end{equation}
At inference, a single Euler step gives $z_1 = z_0 + \hat{v}$, from which the next state is decoded as $\hat{S}_{t+1} = S_t + f_{\text{dec}}(z_1)$. Using more Euler steps ($N{=}2,4$) brought no measurable improvement on our benchmarks.

\textbf{Cross-attention action grounding.}
RLA-WM conditions the dynamics on a pooled action embedding through self-attention, which cannot link a language command to a specific patch. We encode the proposed action $a_t$, a natural-language string such as \texttt{"pick and place red block to the left of blue block"}, with a frozen CLIP text encoder and apply cross-attention with history patch features as queries and CLIP tokens as keys and values. Each history patch $S_{t-h}$ ($h \in \{0, \dots, T_h-1\}$) carries learnable spatial and temporal position embeddings,
\begin{equation}
    \tilde{S}_{t-h, p} = S_{t-h, p} + E_{\text{spat}, p} + E_{\text{temp}, h}.
\end{equation}

\textbf{Causal temporal history.}
Markovian latent models suffer from ``amnesia'' under occlusion: when a gripper encloses an object, most of its features vanish from the current frame. V2A-WM therefore attends to a sliding window of $T_h$ past frames. Episode-start slots are filled with a learned \texttt{[START]} token rather than repeated first frames, which would cause a zero-motion bias, and a binary validity mask restricts attention to valid steps. Finally, an auxiliary sparsity penalty $\mathcal{L}_{\text{sparse}}$ regularizes the velocity field toward zero on static background patches, giving $\mathcal{L} = \mathcal{L}_{\text{FM}} + \lambda \mathcal{L}_{\text{sparse}}$.

\subsection{Dual-Head Critic}
\label{sec:method_critic}
The critic evaluates each imagined trajectory $\{\hat{S}_1, \dots, \hat{S}_H\}$ with two heads:
\begin{align}
    s_{\text{gp}} &= P_{\phi_{\text{gp}}}\big(g \mid \hat{S}_H\big) \label{eq:gp} \\
    s_{\text{tc}} &= \tfrac{1}{H}\textstyle\sum_{h=1}^{H} \cos\left(\phi_{\text{tc}}(\hat{S}_h),\; \phi_{\text{tc}}(\hat{S}_{h-1})\right) \label{eq:tc}
\end{align}

\textbf{Temporal consistency head.} ($s_{\text{tc}}$) is trained contrastively on consecutive versus non-consecutive frame pairs. It also outputs a log-variance, and rollouts whose uncertainty exceeds $\theta_{\text{conf}}$ are discarded. A low $s_{\text{gp}}$ triggers VLM reflection for a revised proposal, and a low $s_{\text{tc}}$ triggers world-model resampling. The selected plan maximizes
\begin{equation}
    \mathcal{C}(\boldsymbol{a}_{1:H}) = \alpha \cdot s_{\text{gp}} + (1 - \alpha) \cdot s_{\text{tc}} .
\end{equation}

\textbf{Spatial goal head.} Spatial goals require patch-level layout, which pooled embeddings do not preserve; e.g., ‘red left of blue’ and ‘blue left of red’ are nearly indistinguishable after pooling. The goal head ($s_{\text{gp}}$) therefore keeps the spatial layout: a 4-layer transformer over the $16{\times}16$ DINOv2 patch tokens and the CLIP text tokens of the goal outputs $P(g \mid \hat{S}_H)$, the probability that the goal holds in the imagined state. We train it with a binary cross-entropy loss on simulator states, whose goal labels follow from the object poses, and on the world model's own imagined states, so that it is calibrated on the kind of input it sees at test time.

\subsection{Sim-to-Real Transfer}
\label{sec:method_twin}
Collecting real transitions for every task is costly, so the world model and the critic that run on the robot are trained only on data generated in a digital twin. The twin is a MuJoCo~\cite{todorov2012mujoco} scene of the workspace, with blocks textured from real camera frames and the camera placed at poses fitted to those frames. The twin is kinematic: since the planner only ever observes the scene with the arm at its home pose, a subtask is rendered as the block moving to its target pose with placement noise, after which physics settles the scene. This yields exactly the $(S_t, a_t, S_{t+1})$ transitions that the world model and critic need, without modeling the arm at all.

\textbf{Real2sim input alignment.} The features of a real frame still differ from those of the training frames. We align the two on the input side: for each planning call, the block poses and the camera pose are fitted to the color masks of the camera frame, and the world model and the critic receive a rendering of the fitted scene in the training domain, while the VLM receives the camera frame itself. If the fit residual exceeds 2.5\,px, verification runs on the camera frame directly. The alignment is not a requirement of the method: Sec.~\ref{sec:real_results} also reports plan ranking on unaligned camera frames, where V2A-WM stays far above chance. Alignment estimates only the current state; candidate outcomes, including invalid preconditions, are still predicted by V2A-WM.

\section{Experiments}
\label{sec:real}

Our experiments on the robot address three questions. (1) Does latent verification improve task success over a VLM alone, and over pixel-space and weaker latent world models, when all methods share the same VLM and the same low-level skills? (2) Do the remaining failures come from the plan or from its execution? (3) What does verification cost?

\begin{figure*}[t]
    \centering
    \includegraphics[width=0.82\textwidth]{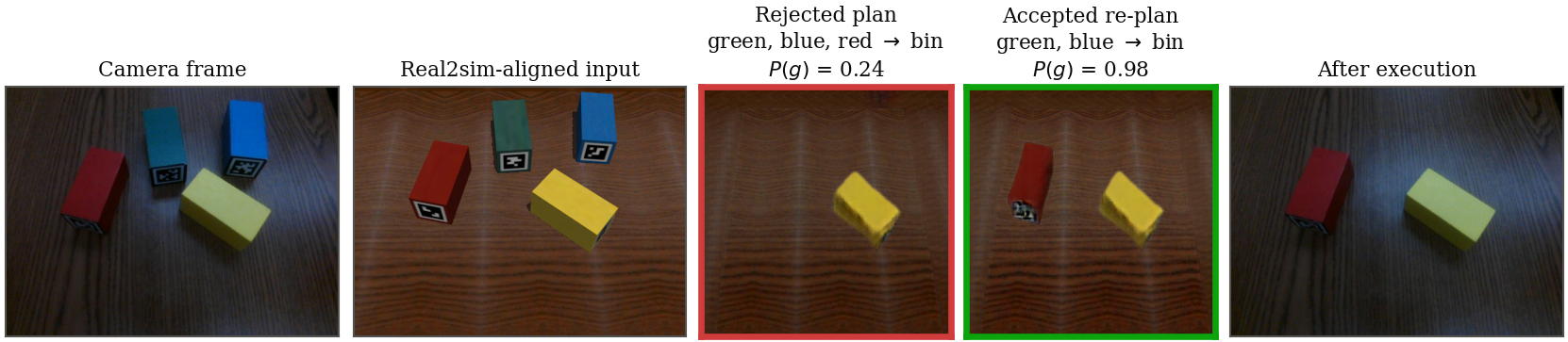}
    \caption{\textbf{One planning call on the robot} (goal: ``Put the red block to the left of the blue block''). After real2sim alignment of the camera frame, V2A-WM imagines each candidate plan and the goal head scores the imagined state. The VLM's first candidate moves the red block, which the world model does not predict to end on the left of the blue block; the second candidate is accepted and executed.}
    \label{fig:real_pipeline}
\end{figure*}

\subsection{Setup}
\label{sec:real_setup}
\textbf{Robot and tasks.} We use a Yahboom DOFBOT with an arm-mounted ($640{\times}480$) RGB camera and four $30{\times}30{\times}60$\ mm colored blocks. The action vocabulary contains three pick-and-place skills:  \texttt{into the bin}, \texttt{on <block>}, and \texttt{to the left/right of <block>}. We evaluate six tasks spanning bin clearing (Bin-2, Bin-cool, Bin-3), spatial relations (Left-of, Right-of), and stacking (Stack).

\textbf{Models.} The twin dataset contains 20k episodes (35k transitions) spanning the three skill families, paraphrased goals, and precondition conflicts. V2A-WM is trained on this data, and the spatial goal head is trained on twin frames together with 22k states imagined by the world model. On held-out twin plans, V2A-WM with its goal head ranks the correct plan first in 96.5\% of cases.

\textbf{Methods.} All methods share the same VLM (GPT-4o), prompts, and low-level skills. Verify2Act uses V2A-WM; RLA-WM replaces it with RLA-WM and a goal head trained on its imagined states; Diffusion-WM follows ReflectVLM~\cite{feng2025reflective} using InstructPix2Pix~\cite{brooks2023instructpix2pix}; VLM-Only executes the first proposal. We use $B{=}3$ candidates, a horizon of $H{=}4$, thresholds $\theta_c{=}\theta_p{=}0.5$, and $\theta_{conf}{=}0.85$, three world-model samples per plan, and at most two re-plans. All methods are trained on the twin dataset. As in simulation, a plan that is still unverified once the re-plan budget is spent is executed anyway.

\textbf{Metrics.} \emph{Task success} is judged by an operator after the plan has been executed. Since all methods share the low-level skills, we also report \emph{plan correctness}, the fraction of executed plans that would satisfy the goal if every skill succeeded, and \emph{step success}, the fraction of executed skills that succeeded. A failed episode counts as a \emph{planning failure} if its plan was incorrect and as an \emph{execution failure} otherwise.

\subsection{Results}
\label{sec:real_results}

\begin{table}[t]
\centering
\caption{\textbf{Task success on the robot (\%).} Avg.: mean over the six tasks for 10 episodes each.}
\label{tab:real_success}
\footnotesize
\setlength{\tabcolsep}{3.2pt}
\begin{tabular}{lccccccc}
\toprule
\textbf{Method} & \rotatebox{60}{Bin-2} & \rotatebox{60}{Bin-cool} & \rotatebox{60}{Bin-3} & \rotatebox{60}{Left-of} & \rotatebox{60}{Right-of} & \rotatebox{60}{Stack} & \textbf{Avg.} \\
\midrule
VLM-Only       & 80 & 40 & 20 & 60 & \textbf{80} & 40 & 53.3 \\
Diffusion-WM   & 80 & 40 & 0  & 40 & \textbf{80} & 40 & 46.7 \\
RLA-WM         & \textbf{100} & 50 & 30 & 80 & 60 & 50 & 61.7 \\
\textbf{Verify2Act (Ours)} & \textbf{100} & \textbf{90} & \textbf{60} & \textbf{90} & \textbf{80} & \textbf{80} & \textbf{83.3} \\
\bottomrule
\end{tabular}
\end{table}

\begin{figure*}[t]
    \centering
    \includegraphics[width=0.84\textwidth]{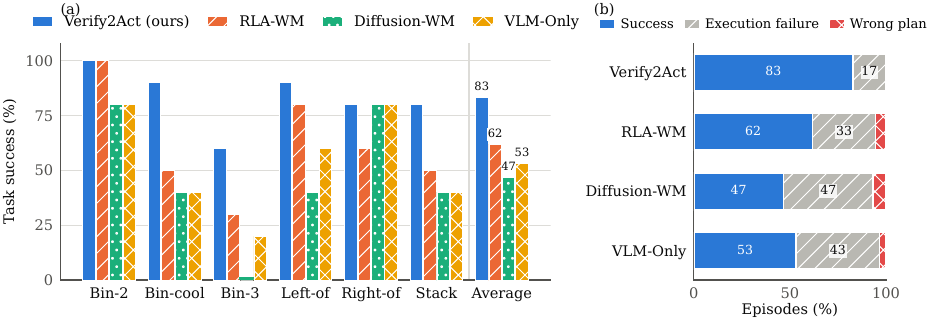}
    \caption{\textbf{Results on the robot.} (a) Task success per task and averaged over tasks. (b) Outcome of the episodes of each method: success, execution failure (the plan was correct but a skill failed), or planning failure (the executed plan could not satisfy the goal).}
    \label{fig:real_results}
\end{figure*}

\begin{table}[t]
\centering
\caption{\textbf{Planning versus execution on the robot.} Plan correct: executed plans that satisfy the goal if all skills succeed. Verified: executed plans accepted by the critic. Top-1: offline plan ranking on 42 real scenes, labeled by simulating every valid plan (chance $\approx$6\%). ``---'': no critic.}
\label{tab:real_planning}
\footnotesize
\setlength{\tabcolsep}{2.8pt}
\begin{tabular}{lcccc}
\toprule
 & \textbf{VLM} & \textbf{Diff.} & \textbf{RLA-} & \textbf{Verify-} \\
 & \textbf{Only} & \textbf{WM} & \textbf{WM} & \textbf{2Act} \\
\midrule
Plan correct (\%)            & 93.3 & 93.3 & 95.0 & \textbf{100} \\
Planning failures (\%)       & 6.7 & 6.7 & 5.0 & \textbf{0.0} \\
Execution failures (\%)      & 43.3 & 46.7 & 33.3 & \textbf{16.7} \\
Step success (\%)            & 69.2 & 61.5 & 73.0 & \textbf{82.5} \\
\midrule
Verified plans (\%)          & --- & --- & 61.8 & \textbf{100} \\
Top-1, real2sim-aligned (\%) & --- & --- & 71.4 & \textbf{97.6} \\
Top-1, camera frame (\%)     & --- & --- & 23.8 & \textbf{71.4} \\
\midrule
VLM calls / episode          & \textbf{1.00} & 2.00 & 1.84 & 1.13 \\
Planning time / call (s)     & \textbf{3.4} & 30.0 & 28.4 & 24.0 \\
Episode time (s)             & \textbf{81} & 107 & 126 & 109 \\
\bottomrule
\end{tabular}
\end{table}

\textbf{Task success.} Table~\ref{tab:real_success} and Fig.~\ref{fig:real_results}(a) give the main result. Averaged over tasks, Verify2Act has the highest success (50/60 = 83.3\%), followed by RLA-WM (37/60 = 61.7\%), VLM-Only (32/60 = 53.3\%) and Diffusion-WM (28/60 = 46.7\%), and it is the best or tied for best on all of the six tasks. We also look beyond mere task success (as shown in Table \ref{tab:real_planning}), which could be influenced by execution failures.

\textbf{Planning versus execution.} Task success mixes planning with low-level execution, so Table~\ref{tab:real_planning} and Fig.~\ref{fig:real_results}(b) separate the two. Two observations put the success rates in context. First, the VLM alone already proposes a correct plan in 93\% of episodes on these tasks, which leaves a verifier little room to improve the plans. Verify2Act nevertheless raises plan correctness to 100\% and is the only method without a planning-induced failure, whereas the others execute an incorrect plan in 5--7\% of episodes.

\textbf{Verification quality.} The effect of the world model shows more clearly in what the critic accepts. Verify2Act verified every plan it executed. With RLA-WM as the world model, 38\% of the plans could not be verified within the re-plan budget and were executed unverified, and the critic's verdict said little about the outcome: unverified plans succeeded about as often (67\%) as verified ones (59\%). To measure verification free of execution noise, we label 42 real scenes by simulating every valid plan from the fitted state, and ask each model to rank the plans (Table~\ref{tab:real_planning}). On the unaligned camera frames, with no real2sim step at all, V2A-WM with its goal head still ranks a correct plan first in 71\% of scenes, against 24\% for RLA-WM and 6\% for chance, so its foresight transfers to real images directly. With real2sim alignment, the rates rise to 97.6\% and 71.4\%: the alignment helps both models, and V2A-WM depends on it far less. Fig.~\ref{fig:real_pipeline} shows a planning call in which the critic rejects the VLM's first candidate on the basis of its imagined outcome and accepts the second. Diffusion-WM often produces unreliable imagined images, so its reflective VLM does not improve over VLM-Only.

\textbf{Cost.} Verify2Act adds 24 s per planning call versus 3.4 s for VLM-Only, while using fewer VLM calls per episode (1.13) than RLA-WM (1.84) or Diffusion-WM (2.00).

\section{Simulation Experiments}
\label{sec:experiments}

We complement the robot experiments with simulation, where longer horizons, more baselines and architectural ablations are practical.

\subsection{Setup}
\label{sec:exp_setup}
We evaluate \textbf{CALVIN}~\cite{mees2022calvin} (\texttt{ABCD$\rightarrow$D}) on 100 five-subtask sequences with HULC~\cite{mees2022hulc}, and Robosuite Cluttered Nut Assembly~\cite{zhu2020robosuite} on 100 episodes with a scripted Cartesian controller. V2A-WM uses frozen DINOv2 ViT-L/14, Adam ($lr{=}10^{-4}$, $T_h{=}3$), 100k training steps, $B{=}3$, $H{=}5$ for CALVIN and $H{=}3$ for Nut Assembly, and $\theta_{\mathrm{conf}}{=}0.85$. Baselines include VLM-Only, DINO-WM (CEM planning, no VLM), RLA-WM, Diffusion-WM, published CALVIN methods, and the V2A-WM (AdaLN) (which replaces cross-attention with AdaLN-style conditioning) and V2A-WM (Non-Causal) (which removes the temporal history window). All methods share the low-level policy of each benchmark.

\begin{table*}[t]
\centering
\caption{\textbf{CALVIN \texttt{ABCD$\rightarrow$D} results}. SR$_k$: fraction of sequences completing at least $k$ consecutive subtasks; \textbf{Avg.~Len.}: mean subtasks completed (max 5); \textbf{Refl.~Calls}: reflective VLM queries triggered by critic rejections. Results are averaged over 3 seeds (mean std. $\pm 1.8\%$). ``---'' denotes not applicable.}
\label{tab:calvin}
\footnotesize
\setlength{\tabcolsep}{4.0pt}
\begin{tabular}{lccccccc}
\toprule
\textbf{Method} &
  \textbf{SR$_1$\,\%} & \textbf{SR$_2$\,\%} & \textbf{SR$_3$\,\%} & \textbf{SR$_4$\,\%} & \textbf{SR$_5$\,\%} &
  \textbf{Avg.~Len.} & \textbf{Refl.~Calls} \\
\midrule
MCIL~\cite{lynch2020language}              & 37.3 & 2.7  & 0.4  & 0.0  & 0.0  & 0.38 & --- \\
HULC~\cite{mees2022hulc}                   & 88.9 & 73.3 & 58.7 & 47.5 & 38.3 & 3.06 & --- \\
GR-1~\cite{wu2023gr1}                      & 94.9 & 89.6 & 84.4 & 78.9 & 73.1 & 4.21 & --- \\
\midrule
VLM-Only (GPT-4o)                          & \textbf{100.0} & 96.0 & \textbf{91.0} & 83.0 & 70.0 & 4.40 & 0 \\
\midrule
DINO-WM~\cite{zhou2024dino}              & 97.0 & 93.0 & 87.0 & 76.0 & 66.0 & 4.19 & 0 \\
RLA-WM~\cite{zhang2026learning}            & \textbf{100.0} & 95.0 & 90.0 & 81.0 & 69.0 & 4.35 & 16 \\
Diffusion-WM~\cite{feng2025reflective}     & \textbf{100.0} & 96.0 & \textbf{91.0} & 79.0 & 66.0 & 4.32 & 444 \\
\midrule
V2A-WM (AdaLN)                             & \textbf{100.0} & 96.0 & 89.0 & 81.0 & 71.0 & 4.37 & 27 \\
V2A-WM (Non-Causal)                        & 99.0 & 94.0 & 90.0 & \textbf{84.0} & 73.0 & 4.40 & 33 \\
\midrule
\textbf{V2A-WM / Verify2Act (Ours)}        & \textbf{100.0} & \textbf{98.0} & 88.0 & 82.0 & \textbf{75.0} & \textbf{4.43} & \textbf{15} \\
\bottomrule
\end{tabular}
\end{table*}

\begin{table}[t]
\centering
\caption{\textbf{CLUTTERED NUT ASSEMBLY RESULTS.} SR: episode success; NCR: nut completion rate; Prec.: critic-accepted plan success; Rej.: critic rejection rate; Repl.: re-plans/episode. Results are averaged over 3 seeds (mean std. $\pm 2.1\%$).}

\label{tab:nutassembly}
\footnotesize
\setlength{\tabcolsep}{3pt}
\begin{tabular}{lccccc}
\toprule
\textbf{Method} & \textbf{SR} & \textbf{NCR} & \textbf{Prec.} & \textbf{Rej.} & \textbf{Repl.} \\
\midrule
VLM-Only (GPT-4o)            & 17.0 & 37.9 & --- & --- & 0.0 \\
DINO-WM~\cite{zhou2024dino}        & 20.0 & 36.9 & 46.6 & 9.2  & 1.5 \\
RLA-WM~\cite{zhang2026learning}      & 43.0 & 67.6 & 76.2 & 69.8 & 17.6 \\
Diffusion-WM~\cite{feng2025reflective} & 55.0 & 69.1 & \textbf{78.9} & 51.8 & 9.0 \\
\midrule
V2A-WM (AdaLN)               & 30.0 & 47.3 & 55.1 & 55.0 & 7.1 \\
V2A-WM (Non-Causal)          & 31.0 & 51.1 & 58.1 & 49.1 & 5.4 \\
\textbf{Verify2Act (Ours)} & \textbf{58.0} & \textbf{70.3} & 61.8 & 49.0 & 10.3 \\
\bottomrule
\end{tabular}
\end{table}

\begin{figure}[t]
  \centering
  \includegraphics[width=0.8\linewidth]{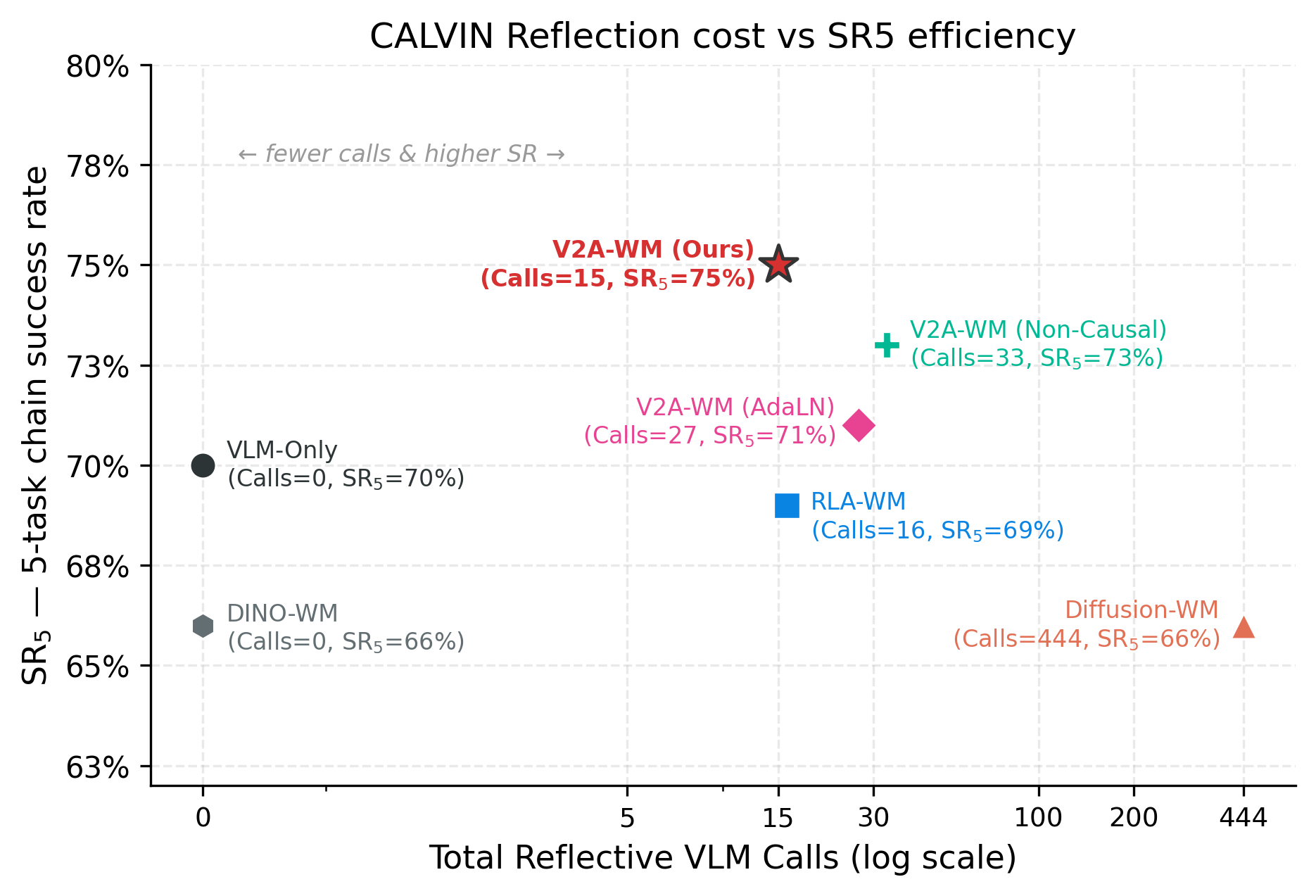}\hfill
  \includegraphics[width=0.8\linewidth]{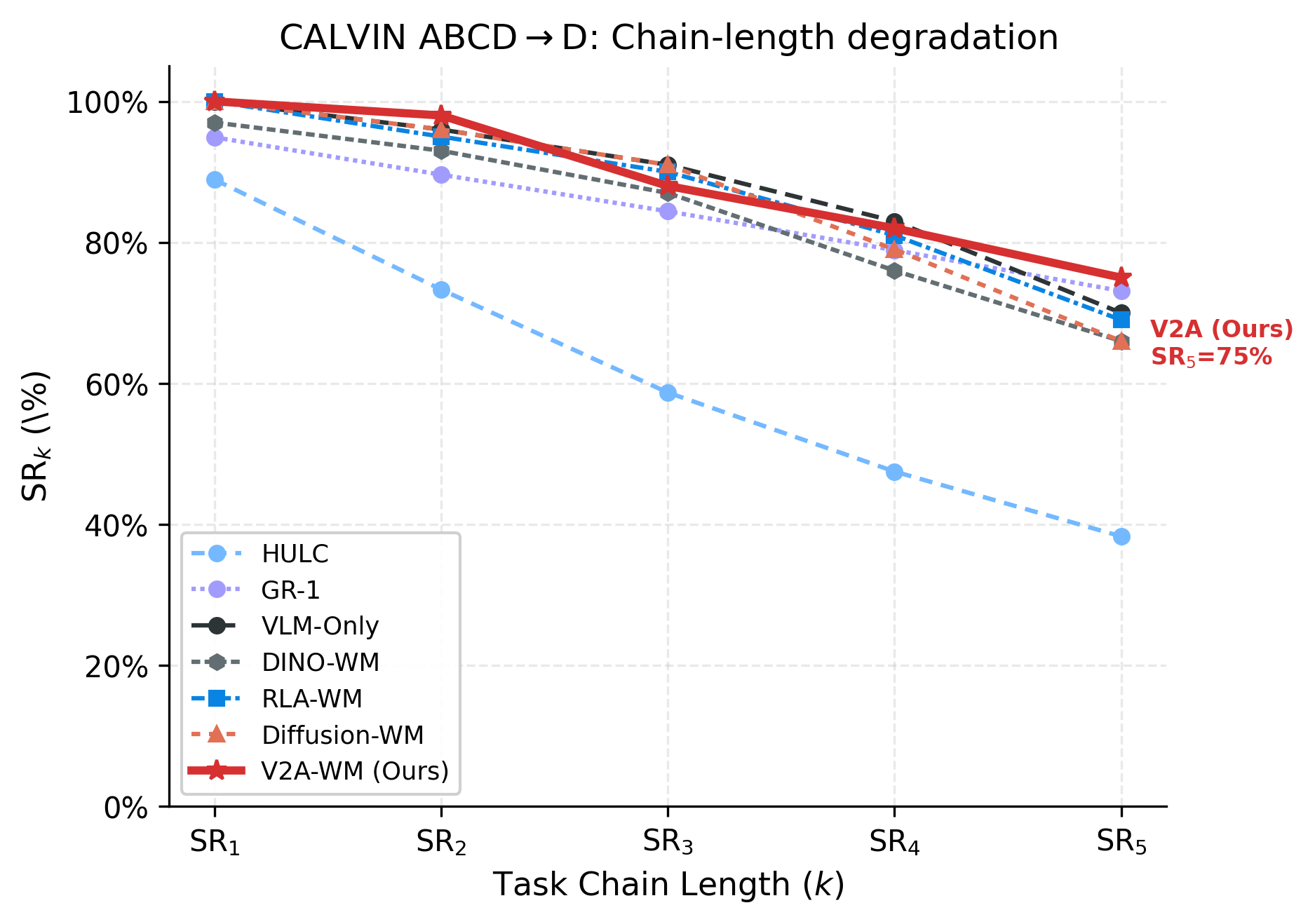}
  \caption{\textbf{CALVIN.} Left: SR$_5$ against reflective VLM calls (log scale). Right: success SR$_k$ over the chain length.}
  \label{fig:calvin_efficiency_combined}
\end{figure}

\subsection{Results}
\label{sec:exp_results}
\textbf{CALVIN.} Verify2Act reaches 75\% SR5, compared with 73.1\% for GR-1 and 70\% for VLM-Only, while triggering only 15 reflective calls versus 444 for Diffusion-WM (i.e 30$\times$ more than Verify2Act). DINO-WM reaches 66\%, and the ablations reach 71\% (AdaLN) and 73\% (Non-Causal). Because the SR5 differences correspond to only a few sequences, we treat these results as supporting rather than conclusive evidence.

\textbf{Nut Assembly.} Verify2Act reaches \textbf{58.0\%} SR and \textbf{70.3\%} NCR (Table~\ref{tab:nutassembly}), above VLM-Only (17.0\%), DINO-WM (20.0\%), RLA-WM (43.0\%) and Diffusion-WM (55.0\%). DINO-WM performs poorly here, where CEM-based planning must search over cluttered multi-nut configurations without language-guided proposals. RLA-WM's critic rejects 69.8\% of plans, which exhausts the re-plan budget on harder episodes, the same over-rejection we observe on the robot. The ablations (30.0\% and 31.0\% SR) show that both cross-attention grounding and temporal history are needed here; critic precision follows the same order (55.1\%, 58.1\%, 61.8\%). Diffusion-WM has the highest critic precision (78.9\%) at a much higher computational cost: a world-model rollout takes 5.1\,s against 0.08\,s for V2A-WM.

\section{Discussion and Limitations}
\label{sec:discussion_limitations}
On the robot, latent verification with V2A-WM achieves the highest task success (83.3\%) and is the only method with no planning-induced failures. It also verifies every executed plan, whereas RLA-WM leaves 38\% of executed plans unverified. Despite being trained entirely on digital-twin data, V2A-WM transfers to the robot with optional real2sim alignment; even without alignment, it ranks the correct plan first in 71\% of real scenes, well above the 6\% chance level.

The results are particularly notable because the VLM is already a strong semantic planner in these tasks: VLM-Only produces a correct plan in 93.3\% of robot episodes. Thus, the benefit of verification is not primarily from correcting a generally poor planner, but from reliably selecting among candidate plans and filtering out the relatively small fraction of incorrect plans before execution. This is reflected in the increase in plan correctness from 93.3\% for VLM-Only to 100\% for Verify2Act and the reduction of planning failures to 0\%. The remaining failures are predominantly due to the shared low-level skill policy rather than high-level planning.

The evaluation has several limitations. Most remaining failures arise from the shared low-level skill policy, leaving limited room for verification on these tasks; scenes with more frequent precondition conflicts would provide a stronger test. Second, the number of real-world trials is modest, and the physical evaluation is restricted to tabletop block manipulation and three primitive skill families. Broader object geometries, articulated objects, and contact-rich manipulation remain to be evaluated. Third, real2sim alignment is restricted to blocks lying on the table and adds several seconds of CPU processing per frame. Finally, the heuristic background-sparsity mask may degrade under dynamic backgrounds or moving cameras, where learned segmentation could be more robust.

\bibliographystyle{IEEEtran}
\bibliography{refs}

@inproceedings{ahn2022can,
  title={Do As {I} Can, Not As {I} Say: Grounding Language in Robotic Affordances},
  author={Ahn, Michael and Brohan, Anthony and Brown, Noah and Chebotar, Yevgen and Cortes, Omar and David, Byron and Finn, Chelsea and Fu, Chuyuan and Gopalakrishnan, Keerthana and Hausman, Karol and others},
  booktitle={Conference on Robot Learning (CoRL)},
  pages={287--318},
  year={2022},
  organization={PMLR}
}

@inproceedings{driess2023palm,
  title={{PaLM-E}: An Embodied Multimodal Language Model},
  author={Driess, Danny and Xia, Fei and Sajjadi, Mehdi SM and Lynch, Corey and Chowdhery, Aakanksha and Ichter, Brian and Wahid, Ayzaan and Tompson, Jonathan and Vuong, Quan and Yu, Tianhe and others},
  booktitle={International Conference on Machine Learning (ICML)},
  pages={8469--8488},
  year={2023},
  organization={PMLR}
}

@inproceedings{brohan2023rt,
  title={{RT-2}: Vision-Language-Action Models Transfer Web Knowledge to Robotic Control},
  author={Brohan, Anthony and Brown, Noah and Carbajal, Justice and Chebotar, Yevgen and Chen, Xi and Choromanski, Krzysztof and Ding, Tianli and Driess, Danny and Dubey, Avinava and Finn, Chelsea and others},
  booktitle={Conference on Robot Learning (CoRL)},
  year={2023},
  organization={PMLR}
}

@inproceedings{feng2025reflective,
  title={Reflective Planning: Vision-Language Models for Multi-Stage Long-Horizon Robotic Manipulation},
  author={Feng, Yunhai and Han, Jiaming and Yang, Zhuoran and Yue, Xiangyu and Levine, Sergey and Luo, Jianlan},
  booktitle={Conference on Robot Learning (CoRL)},
  year={2025},
  organization={PMLR}
}

@inproceedings{black2024zero,
  title={Zero-Shot Robotic Manipulation with Pretrained Image-Editing Diffusion Models},
  author={Black, Kevin and Nakamoto, Mitsuhiko and Atreya, Pranav and Walke, Homer and Finn, Chelsea and Kumar, Aviral and Levine, Sergey},
  booktitle={International Conference on Learning Representations (ICLR)},
  year={2024}
}

@article{zhang2026learning,
  title={Learning Visual Feature-Based World Models via Residual Latent Action},
  author={Zhang, Xinyu and Xu, Zhengtong and Tao, Yutian and Wang, Yeping and She, Yu and Boularias, Abdeslam},
  journal={arXiv preprint arXiv:2605.07079},
  year={2026}
}

@article{hou2026world,
  title={World Model for Robot Learning: A Comprehensive Survey},
  author={Hou, Bohan and Li, Gen and Jia, Jindou and An, Tuo and Guo, Xinying and Leng, Sicong and Geng, Haoran and Ze, Yanjie and Harada, Tatsuya and Torr, Philip and others},
  journal={arXiv preprint arXiv:2605.00080},
  year={2026}
}

@inproceedings{zhou2024dino,
  title={{DINO-WM}: World Models on Pre-trained Visual Features enable Zero-shot Planning},
  author={Zhou, Gaoyue and Pan, Hengkai and LeCun, Yann and Pinto, Lerrel},
  booktitle={International Conference on Machine Learning (ICML)},
  pages={79115--79135},
  year={2025},
  organization={PMLR}
}

@article{oquab2023dinov2,
  title={{DINOv2}: Learning Robust Visual Features without Supervision},
  author={Oquab, Maxime and Darcet, Timoth{\'e}e and Moutakanni, Th{\'e}o and Vo, Huy and Szafraniec, Marc and Khalidov, Vasil and Fernandez, Pierre and Haziza, Daniel and Massa, Francisco and El-Nouby, Alaaeldin and others},
  journal={arXiv preprint arXiv:2304.07193},
  year={2023}
}

@inproceedings{lipman2022flow,
  title={Flow Matching for Generative Modeling},
  author={Lipman, Yaron and Chen, Ricky TQ and Ben-Hamu, Heli and Nickel, Maximilian and Le, Matt},
  booktitle={International Conference on Learning Representations (ICLR)},
  year={2023}
}

@article{mees2022calvin,
  title={{CALVIN}: A Benchmark for Language-Conditioned Policy Learning for Long-Horizon Robot Manipulation Tasks},
  author={Mees, Oier and Hermann, Lukas and Rosete-Beas, Erick and Burgard, Wolfram},
  journal={IEEE Robotics and Automation Letters},
  volume={7},
  number={3},
  pages={7327--7334},
  year={2022},
  publisher={IEEE}
}

@article{zhu2020robosuite,
  title={{robosuite}: A Modular Simulation Framework and Benchmark for Robot Learning},
  author={Zhu, Yuke and Wong, Josiah and Mandlekar, Ajay and Mart{\'\i}n-Mart{\'\i}n, Roberto and Joshi, Abhishek and Nasiriany, Soroush and Zhu, Yifeng},
  journal={arXiv preprint arXiv:2009.12293},
  year={2020}
}

@article{hafner2023dreamerv3,
  title={Mastering diverse domains through world models},
  author={Hafner, Danijar and Pasukonis, Jurgis and Ba, Jimmy and Lillicrap, Timothy},
  journal={arXiv preprint arXiv:2301.04104},
  year={2023}
}

@article{yang2023learning,
  title={Learning interactive real-world simulators},
  author={Yang, Sherry and Du, Yilun and Ghasemipour, Kamyar and Tompson, Jonathan and Kaelbling, Leslie and Schuurmans, Dale and Abbeel, Pieter},
  journal={arXiv preprint arXiv:2310.06114},
  year={2023}
}

@article{garrett2020tamp,
  title={Integrated task and motion planning},
  author={Garrett, Caelan Reed and Chitnis, Rohan and Holladay, Rachel and Kim, Beomjoon and Silver, Tom and Kaelbling, Leslie Pack and Lozano-P{\'e}rez, Tom{\'a}s},
  journal={Annual review of control, robotics, and autonomous systems},
  volume={4},
  number={1},
  pages={265--293},
  year={2021},
  publisher={Annual Reviews}
}

@inproceedings{huang2025points2plans,
  title={{Points2Plans}: From Point Clouds to Long-Horizon Plans with Composable Relational Dynamics},
  author={Huang, Yixuan and Agia, Christopher and Wu, Jimmy and Hermans, Tucker and Bohg, Jeannette},
  booktitle={IEEE International Conference on Robotics and Automation (ICRA)},
  year={2025}
}

@inproceedings{radford2021clip,
  title={Learning Transferable Visual Models From Natural Language Supervision},
  author={Radford, Alec and Kim, Jong Wook and Hallacy, Chris and Ramesh, Aditya and Goh, Gabriel and Agarwal, Sandhini and Sastry, Girish and Askell, Amanda and Mishkin, Pamela and Clark, Jack and others},
  booktitle={International Conference on Machine Learning},
  pages={8748--8763},
  year={2021},
  organization={PMLR}
}

@inproceedings{du2023learning,
  title={Learning Universal Policies via Text-Guided Video Generation},
  author={Du, Yilun and Yang, Sherry and Dai, Bo and Dai, Hanjun and Nachum, Ofir and Tenenbaum, Joshua B. and Schuurmans, Dale and Abbeel, Pieter},
  booktitle={Advances in Neural Information Processing Systems (NeurIPS)},
  year={2023}
}

@article{gu2026say,
  title={Say, Dream, and Act: Learning Video World Models for Instruction-Driven Robot Manipulation},
  author={Gu, Songen and Cai, Yunuo and Wang, Tianyu and Wu, Simo and Fu, Yanwei},
  journal={arXiv preprint arXiv:2602.10717},
  year={2026}
}

@article{jia2025video2act,
  title={{Video2Act}: A Dual-System Video Diffusion Policy with Robotic Spatio-Motional Modeling},
  author={Jia, Yueru and Liu, Jiaming and Liu, Shengbang and Zhou, Rui and Yu, Wanhe and Yan, Yuyang and Chi, Xiaowei and Guo, Yandong and Shi, Boxin and Zhang, Shanghang},
  journal={arXiv preprint arXiv:2512.03044},
  year={2025}
}

@article{pai2025mimic,
  title={mimic-video: Video-Action Models for Generalizable Robot Control Beyond {VLAs}},
  author={Pai, Jonas and Achenbach, Liam and Montesinos, Victoriano and Forrai, Benedek and Mees, Oier and Nava, Elvis},
  journal={arXiv preprint arXiv:2512.15692},
  year={2025}
}

@inproceedings{li2025unified,
  title={Unified Video Action Model},
  author={Li, Shuang and Gao, Yihuai and Sadigh, Dorsa and Song, Shuran},
  booktitle={Robotics: Science and Systems (RSS)},
  year={2025}
}

@inproceedings{zhu2025unified,
  title={Unified World Models: Coupling Video and Action Diffusion for Pretraining on Large Robotic Datasets},
  author={Zhu, Chuning and Yu, Raymond and Feng, Siyuan and Burchfiel, Benjamin and Shah, Paarth and Gupta, Abhishek},
  booktitle={Robotics: Science and Systems (RSS)},
  year={2025}
}

@article{kim2026cosmos,
  title={{Cosmos Policy}: Fine-Tuning Video Models for Visuomotor Control and Planning},
  author={Kim, Moo Jin and Gao, Yihuai and Lin, Tsung-Yi and Lin, Yen-Chen and Ge, Yunhao and Lam, Grace and Liang, Percy and Song, Shuran and Liu, Ming-Yu and Finn, Chelsea and others},
  journal={arXiv preprint arXiv:2601.16163},
  year={2026}
}

@article{zheng2025flare,
  title={{FLARE}: Robot Learning with Implicit World Modeling},
  author={Zheng, Ruijie and Wang, Jing and Reed, Scott and Bjorck, Johan and Fang, Yu and Hu, Fengyuan and Jang, Joel and Kundalia, Kaushil and Lin, Zongyu and Magne, Loic and others},
  journal={arXiv preprint arXiv:2505.15659},
  year={2025}
}

@article{sun2026vlajepa,
  title={{VLA-JEPA}: Enhancing Vision-Language-Action Model with Latent World Model},
  author={Sun, Jingwen and Zhang, Wenyao and Qi, Zekun and Ren, Shaojie and Liu, Zezhi and Zhu, Hanxin and Sun, Guangzhong and Jin, Xin and Chen, Zhibo},
  journal={arXiv preprint arXiv:2602.10098},
  year={2026}
}

@article{assran2025vjepa,
  title={{V-JEPA 2}: Self-Supervised Video Models Enable Understanding, Prediction and Planning},
  author={Assran, Mahmoud and Bardes, Adrien and Fan, David and Garrido, Quentin and Howes, Russell and Komeili, Mojtaba and Muckley, Matthew J and Rizvi, Ammar and Roberts, Claire and Sinha, Koustuv and others},
  journal={arXiv preprint arXiv:2506.09985},
  year={2025}
}

@article{miao2026jepavla,
  title={{JEPA-VLA}: Video Predictive Embedding is Needed for {VLA} Models},
  author={Miao, Shangchen and Feng, Ningya and Wu, Jialong and Lin, Ye and He, Xu and Li, Dong and Long, Mingsheng},
  journal={arXiv preprint arXiv:2602.11832},
  year={2026}
}

@article{su2026wog,
  title={World Guidance: World Modeling in Condition Space for Action Generation},
  author={Su, Yue and Chen, Sijin and Shi, Haixin and Liu, Mingyu and Zhang, Zhengshen and Huang, Ningyuan and Zhong, Weiheng and Zhu, Zhengbang and Liu, Yuxiao and Liu, Xihui},
  journal={arXiv preprint arXiv:2602.22010},
  year={2026}
}

@article{chen2026dial,
  title={Dial: Decoupling intent and action via latent world modeling for end-to-end vla},
  author={Chen, Yi and Ge, Yuying and Zhou, Hui and Ding, Mingyu and Ge, Yixiao and Liu, Xihui},
  journal={arXiv preprint arXiv:2603.29844},
  year={2026}
}

@inproceedings{wu2023gr1,
  title={Unleashing Large-Scale Video Generative Pre-training for Visual Robot Manipulation},
  author={Wu, Hongtao and Jing, Ya and Cheang, Chilam and Chen, Guangzeng and Xu, Jiafeng and Li, Xinghang and Liu, Minghuan and Li, Hang and Kong, Tao},
  booktitle={International Conference on Learning Representations (ICLR)},
  year={2024}
}

@article{mees2022hulc,
  title={What Matters in Language Conditioned Robotic Imitation Learning over Unstructured Data},
  author={Mees, Oier and Hermann, Lukas and Burgard, Wolfram},
  journal={IEEE Robotics and Automation Letters},
  volume={7},
  number={4},
  pages={11205--11212},
  year={2022}
}

@inproceedings{lynch2020language,
  title={Language Conditioned Imitation Learning over Unstructured Data},
  author={Lynch, Corey and Sermanet, Pierre},
  booktitle={Robotics: Science and Systems (RSS)},
  year={2021}
}

@inproceedings{brooks2023instructpix2pix,
  title={{InstructPix2Pix}: Learning to Follow Image Editing Instructions},
  author={Brooks, Tim and Holynski, Aleksander and Efros, Alexei A},
  booktitle={IEEE/CVF Conference on Computer Vision and Pattern Recognition (CVPR)},
  pages={18392--18402},
  year={2023}
}

@article{flowwm2024,
  title={Flow Matching in Feature Space for Stochastic World Modeling},
  author={Porcher, Fran{\c{c}}ois and Carion, Nicolas and Alahari, Karteek and Chen, Shizhe},
  journal={arXiv preprint arXiv:2606.29059},
  year={2026}
}

@article{mowm2024,
  title={{MoWM}: Mixture-of-World-Models for Embodied Planning via Latent-to-Pixel Feature Modulation},
  author={Yu, Yangcheng and Jin, Xin and Shang, Yu and Zhang, Xin and Su, Haisheng and Wu, Wei and Li, Yong},
  journal={arXiv preprint arXiv:2509.21797},
  year={2025}
}

@article{flowmpc2024,
  title={{FlowMPC}: Improving Flow Matching policies with World Models},
  author={Hamel, Chandon},
  journal={arXiv preprint arXiv:2606.16286},
  year={2026}
}

@inproceedings{todorov2012mujoco,
  title={{MuJoCo}: A physics engine for model-based control},
  author={Todorov, Emanuel and Erez, Tom and Tassa, Yuval},
  booktitle={IEEE/RSJ International Conference on Intelligent Robots and Systems (IROS)},
  pages={5026--5033},
  year={2012}
}

\end{document}